\documentclass[12pt,letterpaper]{article}

\usepackage[margin=1in]{geometry}
\usepackage[T1]{fontenc}
\usepackage[utf8]{inputenc}
\usepackage{lmodern}
\usepackage{hyperref}
\usepackage{natbib}
\usepackage{amsmath}
\usepackage{amsfonts}
\usepackage{graphicx}
\usepackage{subcaption}
\usepackage{listings}
\usepackage{booktabs} 

\DeclareMathOperator*{\argmin}{arg\,min}

\title{LLMPC: Large Language Model Predictive Control}
\author{Gabriel Maher \thanks{gabriel.d.maher@gmail.com}}
\date{\today}

\begin{document}

\maketitle

\begin{abstract}
Recent advancements in prompting techniques for Large Language Models (LLMs) have improved their reasoning, planning, and action abilities. 
This paper examines these prompting techniques through the lens of model predictive control (MPC). 
We show that LLMs act as implicit planning cost function minimizers when planning prompts are used.
We propose a unified MPC framework for planning with LLMs and demonstrate improved performance over few shot prompting on several planning benchmarks.
\end{abstract}

\section{Introduction}
Recent work has demonstrated remarkable success in enhancing Large Language Model (LLM) capabilities through integration with planning and control algorithms. From early prompting techniques like chain-of-thought \citep{wei2022chainofthought} and self-consistency \citep{wang2022selfconsistency}, to more sophisticated approaches combining LLMs with tree search \citep{wang2024litesearch,jiang2024treesearch} and reinforcement learning \citep{DeepSeekR1_2025}, researchers have found that guiding LLM outputs with structured planning significantly improves performance on complex reasoning tasks.

In robotics and embodied AI, frameworks like SayCan \citep{ahn2022doasican} and Voyager \citep{wang2023voyager} successfully use LLMs to generate action plans for physical systems. LLM-Assist demonstrated improved autonomous vehicle planning by combining LLMs with traditional planners \citep{song2023llmplanner}. Knowledge-guided approaches like KnowAgent \citep{zhu2024knowagent} further enhance LLM planning by incorporating domain expertise and constraints.

Several approaches have explored using LLMs for hierarchical and multi-agent planning. Two-step goal decomposition \citep{singh2024twostepmultiagenttaskplanning} showed that LLMs could break down complex multi-agent tasks more effectively than traditional PDDL planners. ISR-LLM \citep{zhou2024isr} and hierarchical constraint planning \citep{zhang2024planningmulticonstraintscollaborativelanguage} demonstrated improved performance on problems requiring coordination between multiple constraints and objectives.

Search-based methods have proven particularly effective when combined with LLMs. Tree search algorithms \citep{hao2023reasoninglanguagemodelplanning, wang2024litesearch,jiang2024treesearch,putta2024agentq} use LLMs to guide exploration while value functions evaluate candidate solutions. Scattered forest search \citep{light2024scatteredforestsearchsmarter} showed improved solution diversity in code generation tasks. Natural language planning \citep{wang2024planningnaturallanguageimproves} demonstrated that describing search spaces in natural language improved code generation performance.
In the context of web agents, some work has explored the use of LLMs as world models \citep{gu2024llmsecretlyworldmodel,chae2024webagentsworldmodels}.

Recent work has also explored enhancing LLM planning through learning. Supervised fine-tuning approaches like TravelPlanner \citep{chen2024travelplanner} showed improved planning performance through targeted training. World models have been used to enable more structured planning \citep{xiong2024deliberatereasoningllmsstructureaware}, while continual learning approaches \citep{paul2024continualplanning} demonstrated better exploration in reinforcement learning settings.

The success of these hybrid approaches raises important questions about the relationship between LLMs and classical planning algorithms. While various methods have shown empirical benefits, we lack a formal framework for understanding how LLMs function as planners and how they can be optimally integrated with traditional control techniques.
In particular, the connection between LLM planning and Model Predictive Control (MPC) \citep{GARCIA1989335,mayne2000mpc} - a powerful framework for generating and executing control sequences - remains largely unexplored.

This paper bridges this gap by examining LLM planning through the lens of MPC. We show that LLMs inherently act as approximate optimization algorithms when generating plans, and that making this optimization explicit through an MPC framework can significantly improve performance. Our approach provides both theoretical insights into why structured planning works with LLMs and practical methods for enhancing LLM planning capabilities.

The code for our experiments is available at \url{https://github.com/gmaher/llmpc}.

\section{Model Predictive Control}
In the model predictive control setting an agent must navigate a state space $s_t \in S\subset\mathbb{R}^d$. The agent decides on an action $a_t \in A \subset \mathbb{R}^m$.
The state is then updated according to a state transition model $s_{t+1} = f\left(s_t, a_t, \varepsilon_t\right)$, where $\varepsilon_t$ is a noise or disturbance term.

The goal is to find a sequence of actions $a_t, a_{t+1}, \dots, a_{t+H}$ over a planning horizon of $H$ steps that minimizes an objective function $C\left(s_t, \dots, s_{t+H}, a_t, a_{t+1}, \dots, a_{t+H}\right)$.
That is
\begin{equation}
    \hat{a}_t\, \dots\, \hat{a}_{t+H} = g\left(s_t\right) := \argmin_{a_t, a_{t+1}, \dots, a_{t+H}} C\left(s_t, \dots, s_{t+H}, a_t, \dots, a_{t+H}\right)
\end{equation}
The objective function typically involves a task-specific cost and a regularization cost.
The task-specific cost is often a measure of the distance between the current state and a desired state.
The regularization cost is often a measure of the complexity of the sequence of actions.

\section{LLM as MPC Plan Sampler}
An LLM is a neural network function that operates on a length $L$ sequence of input tokens $q_t \in Q^L := \{1, 2, \dots, M\}^L$ to produce a probability vector over the possible $M$ tokens for the next token in the sequence \citep{Paaß2023,NEURIPS2020_1457c0d6}.
The LLM input is typically referred to as a `prompt' and `prompting' is the process through which relevant state information from $s_t$ is embedded into the prompt.
In the control scenario the previous prompt $q_{t-1}$ and the current state $s_t$ create the current prompt $q_t$, thus prompting can be represented as a function $P:\mathcal{S}\times Q^L \to Q^L$. 

Calling the LLM produces a probability distribution for the next token. Various methods are used for sampling the actual next token value \citep{holtzman2020curiouscaseneuraltext}. For example, Beam Search is a common approach where several likely token sequences are sampled by selecting the top most likely tokens at each step and using these to sample subsequent tokens with repeated pruning to remove unlikely sequences.
Repeatedly evaluating the LLM with the new output tokens produces a sequence of output tokens $\hat{Y} \in \{1, 2, \dots, M\}^T$.
We can represent the process of sampling the output token sequence as a function $F:\{1, 2, \dots, M\}^L \to \{1, 2, \dots, M\}^T$.

In the LLM-as-planner scenario, the token sequence is mapped to the sequence of actions $a_t,\dots,a_{t+H}$ for the controller. That is there is a mapping function $\phi : \{1, 2, \dots, M\}^T \to A^H$.
We now see that the LLM-as-planner approximately minimizes the cost function through
\begin{equation}
\hat{a}_t\, \dots\, \hat{a}_{t+H} = \phi\left(\hat{Y}\right) := \phi\left(F(q_t)\right) := \phi\left(F\left(P\left(s_t,q_{t-1}\right)\right)\right) \approx g\left(s_t\right)
\end{equation}

In the MPC framework, the main limitation of LLM planning performance is due to the limits of LLM sampling as approximately minimizing the MPC objective.
Thus by making better use of the objective function we can improve the performance of the LLM planners.
Since the LLM output $\hat{Y}$ is a sequence of random variables, the LLM generated plan will also be random.
We propose using the LLM to sample a number of control sequences and then to use the cost function to evaluate and rank each plan.
The best sample is selected as the control sequence to apply.

That is we sample a number of plans $A^i_t:=\{a^i_t,\dots,a^i_{t+H}\}$ from the LLM.
For each plan we simulate the system to obtain the states $S^i_t:=\{s^i_t,\dots,s^i_{t+H}\}$.
We then pick the plan that minimizes the objective function among the sampled plans
\begin{equation}
    \hat{A}_t := \arg \min_{i=1,\dots,K} C\left(S^i_t, A^i_t\right)
\end{equation}
We then execute up to $H$ steps from the plan $\hat{A}_t$ and then replan with the new state information.
We note that while we use the LLM to sample plans directly, they could also be sampled one action at a time while using beam-search or other algorithms to construct the final plan samples.

\section{Experiments}

\subsection{Control of Spring and Mass System}
Here we compare LLMPC against MPC on the problem of applying force to a spring and mass system to arrive at a particular goal state.
The equations of motion for the system are
\begin{align*}
a_t &= \frac{1}{m}\left(u - k(x-x_0-l)\right), \\
v_{t+1} &=  v_t + \mathrm{d}t a_t, \\
x_{t+1} &= x_t + \mathrm{d}t v_t.
\end{align*}

Here $u$ is a force applied by the controller to control the spring.
The objective is to bring the spring to the goal state $x^*$ with zero velocity.
The objective function is
\begin{equation}
    c_t = Q_x\left(x_{t+H}-x^*\right)^2 + Q_v\left(v_{t+H}-v^*\right) + \sum_{k=t}^{t+H}Q_u u_k^2
\label{eq:springcost}
\end{equation}

For both MPC and LLMPC we set $H=3$ and execute 2 of every 3 steps from the returned action sequence.
For MPC we use CVXPY to solve the planning problem.
For LLMPC we used GPT-4o-mini and asked the LLM to suggest 5 plans at every step using a templated prompt (listing \ref{lst:springprompt}).
Each of the suggested action sequences is evaluated using \eqref{eq:springcost}, and the plan with the lowest objective value is selected.

\begin{figure}[h]
    \centering
    
    \begin{subfigure}[b]{0.9\textwidth}
        \centering
        \includegraphics[width=\textwidth]{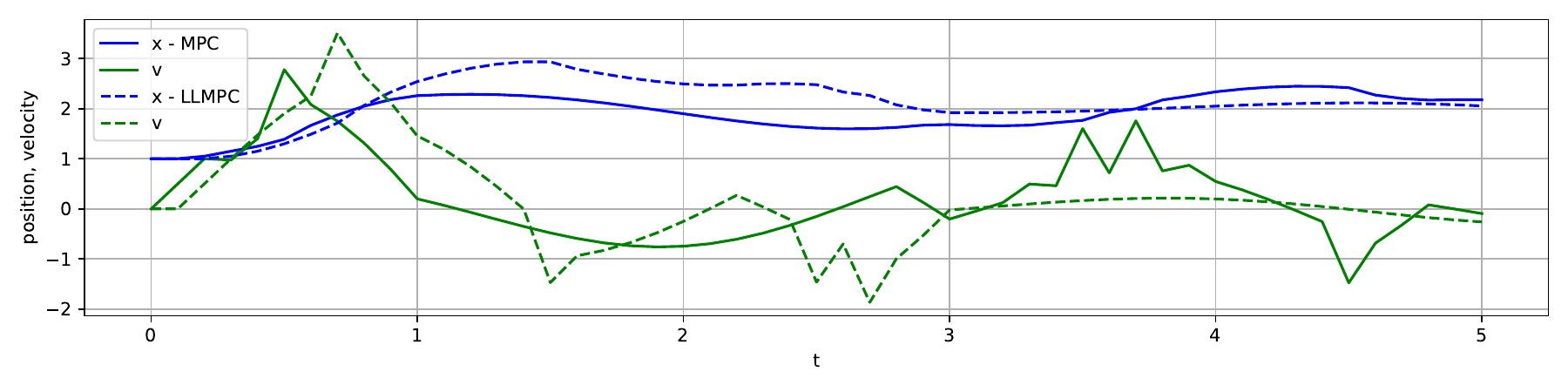}
        \caption{Position and velocity trajectory}
        \label{fig:position}
    \end{subfigure}
    
    \begin{subfigure}[b]{0.9\textwidth}
        \centering
        \includegraphics[width=\textwidth]{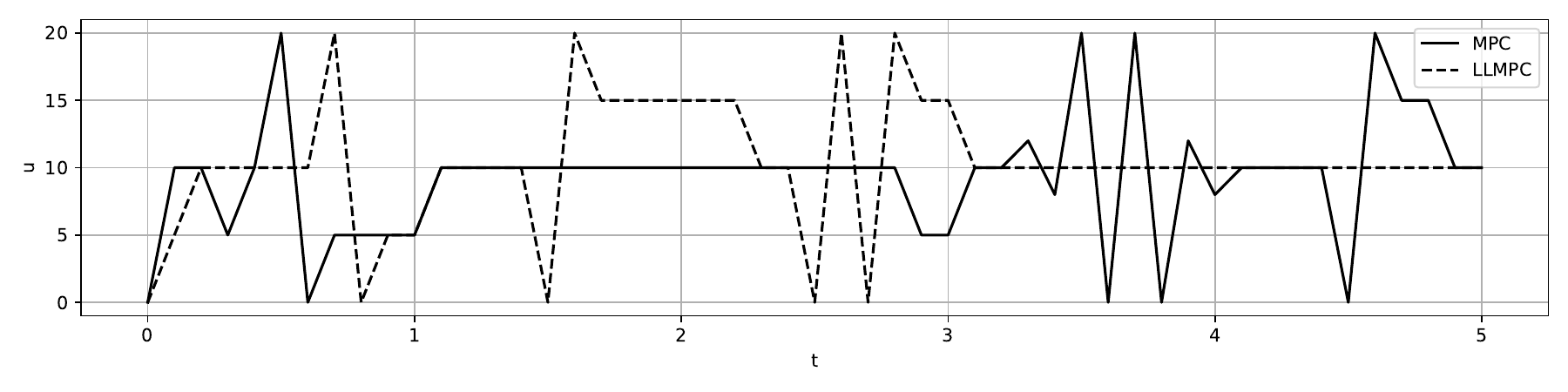}
        \caption{Control trajectory}
        \label{fig:control}
    \end{subfigure}
    
    \begin{subfigure}[b]{0.9\textwidth}
        \centering
        \includegraphics[width=\textwidth]{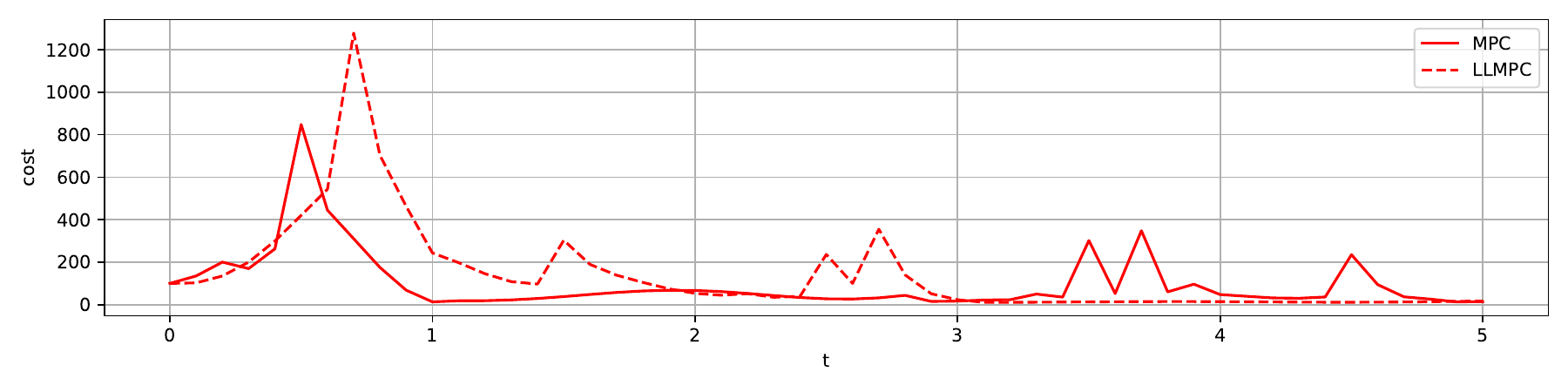}
        \caption{Objective trajectory}
        \label{fig:objective}
    \end{subfigure}

    \caption{State, control and cost trajectories for MPC and LLMPC algorithm on spring-mass problem.}
    \label{fig:spring}
\end{figure}

We solved the control problem with both MPC and LLMPC with $x_0=1$, $v_0=0$, $x^*=2$, $v^*=0$, $m=1$, and $k=5$ (Fig. \ref{fig:spring}).
Both MPC and LLMPC produce control sequences that control the spring to the goal state.
As expected, the objective values from the plans produced by LLMPC are higher than when solving the problem exactly with MPC, highlighting that LLMs are approximate planners.

\begin{figure}
    \centering
    \includegraphics[width=0.8\textwidth]{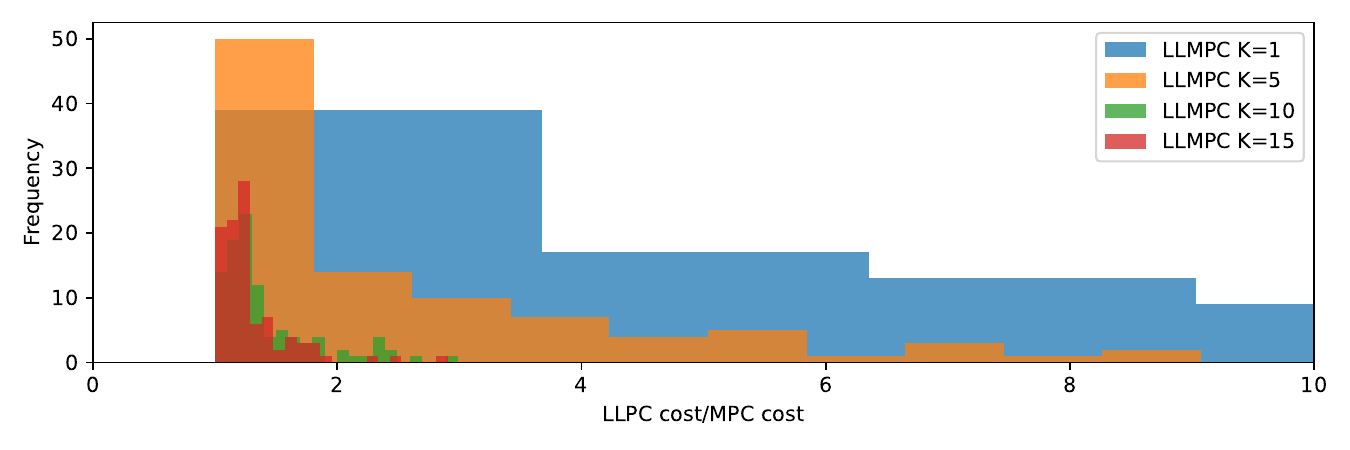}
    \caption{Ratio of LLMPC optimized planning objective value to MPC optimized planning objective value for mass-spring system over a range of states.}
    \label{fig:spring_compare_bar}
\end{figure}

To further illustrate the approximate planning capacties of LLMs we sample a range of state vectors $s_t$ for the mass-spring system and solve the planning objective using both LLMPC and MPC at each sampled state.
We then calculate the ratio of the optimized objective value returned by LLMPC against that of MPC (Tab. \ref{tab:spring_compare}, Fig. \ref{fig:spring_compare_bar}). We see that sampling only a single plan from an LLM produces poor objective values when compared to MPC (objective ratio of 8.21).
However as we increase the number of plans the performance of LLMPC significantly increases with the ratio decreasing to 1.3 when 15 plans are sampled per iteration. 

\begin{table}[ht]
    \centering
    \caption{Mean objective ratio for LLMPC vs MPC across range of states and number of plans sampled}
    \begin{tabular}{lcccc}
        \toprule
         & $K=1$ & $K=5$ & $K=10$ & $K=15$ \\
        \midrule
        Ratio LLMPC cost/MPC cost & 8.21 & 2.89 & 1.42 & 1.30 \\
        \bottomrule
    \end{tabular}
    \label{tab:spring_compare}
\end{table}

\subsection{Trip Planning}
We evaluate LLMPC on the trip-planning test case from the Natural Plan benchmark \citep{zheng2024naturalplan}.
The trip planning problem involves proposing a travel itinerary to visit multiple cities with given constraints on the number of days to visit each city and the available flights between cities (see listing \ref{lst:trip_example} for an example).
We compare LLMPC with increasing iteration budgets against a single round few-shot prompt solution.

We construct a sequential planning system prompt and instruction prompt (Listings \ref{lst:trip_system}, \ref{lst:trip_instruction}) that provides the problem description to an LLM along with few-shot examples and feedback on the previously proposed plan.
To provide feedback on the previous plan we use an evaluation function that checks whether the constraints from the problem description are met and lists any unmet constraints.
If no plan has been proposed yet the feedback section is omitted.
The LLM is asked to then propose a new plan to address any issue and solve the problem.
With these prompts the trip planning problem can be solved using LLMPC by proposing an initial plan without feedback and then sequentially iterating and improving the proposed plan. 

\begin{figure}[ht]
    \centering
    \includegraphics[width=0.8\textwidth]{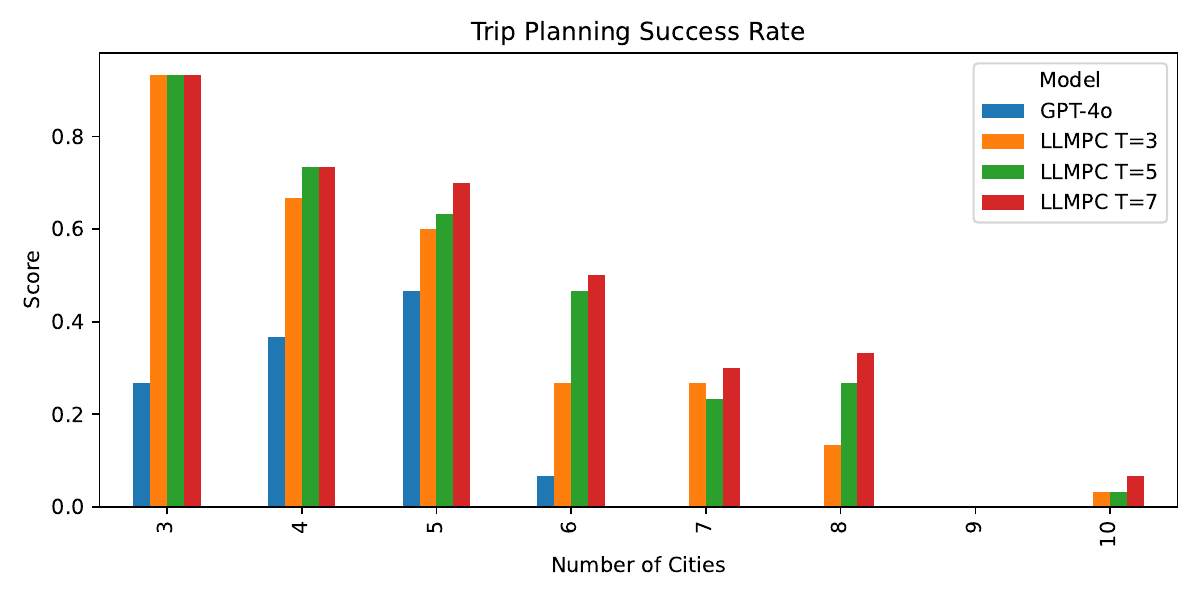}
    \caption{Trip planning success rate grouped by number of cities}
    \label{fig:trip_planning_bar}
\end{figure}

We measure the overall success rate and success rate segmented by the number of cities the problem description lists.
Adding more cities to the problem increases its complexity and difficulty.
The results show that LLMPC significantly outperforms single-round GPT-4o planning, with success rates improving from 14.5\% to 44.6\% as we increase the number of iterations (Table \ref{tab:trip-planning}). The performance gains are particularly notable for problems with more cities to visit (Figure \ref{fig:trip_planning_bar}). Increasing the iteration budget for LLMPC improves the performance as problem complexity increases.

\begin{table}[ht]
    \centering
    \caption{Trip Planning Success Rate for single round GPT-4o and LLMPC}
    \begin{tabular}{lcccc}
        \toprule
         & \textbf{GPT-4o} & \textbf{LLMPC} $T=3$ & \textbf{LLMPC} $T=5$ & \textbf{LLMPC} $T=7$ \\
        \midrule
        Success Rate & 0.145 & 0.363 & 0.413 & \textbf{0.446} \\
        \bottomrule
    \end{tabular}
    \label{tab:trip-planning}
\end{table}

\subsection{Meeting Planning}
We evaluate LLMPC on the meeting planning test case from the Natural Plan benchmark \citep{zheng2024naturalplan}. In this problem, a user must plan meetings with multiple friends in different locations across a city while satisfying temporal and spatial constraints. Each friend is available at a specific location during a fixed time window and requires a minimum meeting duration. The planner must account for travel times between locations, meeting duration requirements, and friend availability windows to construct a valid schedule that maximizes the number of successful meetings.

The meeting planning problem presents several key challenges that make it an ideal test case for LLMPC.
There are complex temporal dependencies between travel times, meeting durations, and availability windows. 
Additionally the problem contains spatial constraints from the city travel network that limit possible meeting sequences. 
Finally the problem involves multiple competing objectives around maximizing meetings while satisfying minimum durations

For LLMPC we again construct a sytem prompt and instruction prompt that provides the problem description, few-shot examples and plan feedback as input (Listings \ref{lst:meeting_system}, \ref{lst:meeting_instruction}).
We again use an evaluation function to assess any unmet constraints of the proposed plan and provide these as feedback.

We compare LLMPC against single-round few-shot prompting and evaluate different configurations:
\begin{enumerate}
    \item Varying the number of iterations T to refine plans
    \item Sampling multiple plans per iteration (K > 1)
    \item Combinations of iteration and sampling to balance exploration and refinement    
\end{enumerate}
We measure overall success rate and success rate segmented by number of friends that are listed in the problem description.
More meetings increases the complexity and difficulty of the problem.

\begin{figure}[ht]
    \centering
\includegraphics[width=\textwidth]{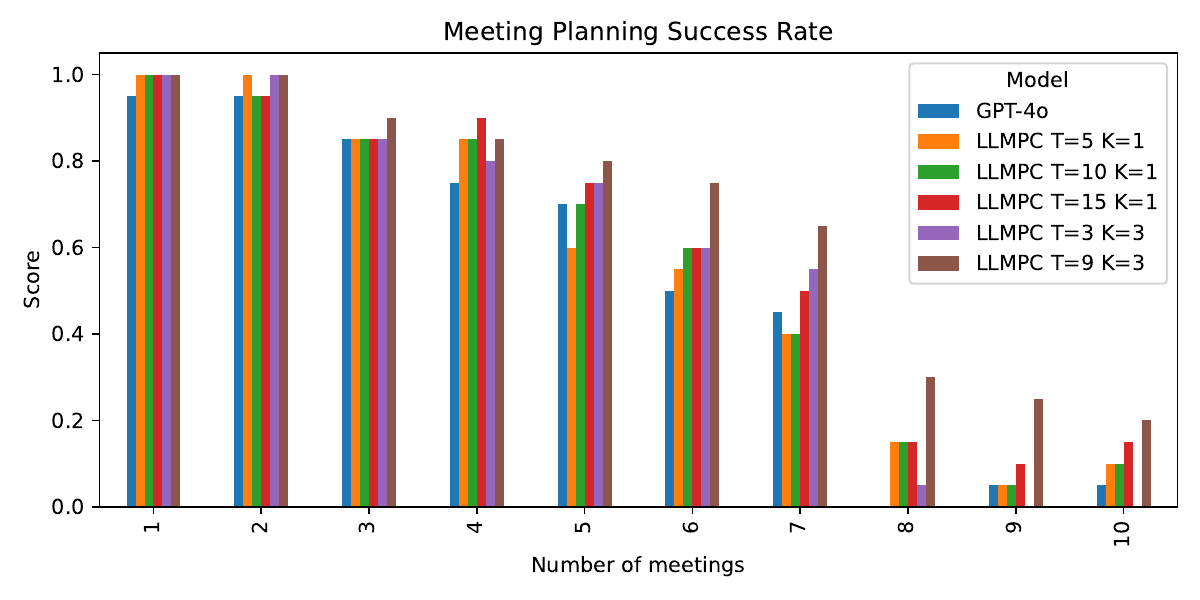}
\caption{Comparison of LLMPC against GPT-4o with few shot prompt on meeting planning problem, success rate segmented by number of meetings.}
\label{fig:meeting_planning_bar}
\end{figure}

The results demonstrate that LLMPC provides consistent improvements over single-round planning (Table \ref{tab:meeting-planning}). Increasing both the number of iterations $T$ and plans per iteration $K$ leads to better performance, with the best results achieved using $T=9$ and $K=3$ (67\% success rate vs 52.5\% for single-round planning). These results highlight the fact that the largest improvements come from both sampling multiple plans and iterating on previous solutions. Figure \ref{fig:meeting_planning_bar} shows the performance breakdown, highlighting LLMPC's ability to better handle the temporal and spatial constraints of the meeting planning problem. For the meeting planning problem as complexity increases it becomes increasingly necessary to both iterate and sample multiple plans.

\begin{table}[ht]
    \centering
    \caption{Meeting Planning Success Rate for single round GPT-4o and LLMPC}
    \begin{tabular}{lcccccc}
        \toprule
         & \textbf{GPT-4o} & \textbf{LLMPC} & \textbf{LLMPC}  & \textbf{LLMPC}  & \textbf{LLMPC}  & \textbf{LLMPC}  \\
         & &$T=5$ &$T=10$ &$T=15$ &$T=3, K=3$ &$T=9, K=3$ \\
        \midrule
        Success Rate & 0.525 & 0.555 & 0.565 & 0.595 & 0.56 & \textbf{0.67}\\
        \bottomrule
    \end{tabular}
    \label{tab:meeting-planning}
\end{table}

\section{Conclusion}
This paper introduced LLMPC - a framework for structured planning with large language models based on model predictive control principles. We showed that LLMs naturally act as implicit optimization algorithms when generating plans and that making this optimization explicit through MPC improves performance. 

A key insight of our work are that LLMs can be viewed as approximate optimizers of planning cost functions. Furthermore sampling multiple plans from LLMs and selecting the best according to an objective function significantly improves planning performance. Finally iterative refinement through replanning with updated state information further enhances solution quality.

We demonstrated these benefits empirically on three test cases - a mass-spring control problem, trip planning, and meeting scheduling. In all cases, LLMPC improved performance over single-round few-shot prompting, with the gains increasing as problem complexity grows. For the spring-mass system, increasing the number of sampled plans reduced the objective gap with exact MPC from 8.21x to 1.3x. On the Natural Plan benchmarks, LLMPC improved success rates by 30.1\% on trip planning and 14.5\% on meeting scheduling.

Future work could explore several promising directions. Combining LLMPC with other planning techniques like beam search may yield further improvements. Additional work could explore the use of learned cost functions and state update functions when exact versions are not available.

The LLMPC framework provides both practical benefits for real-world planning problems and theoretical insights into how LLMs function as planners. We believe this perspective will be valuable for developing more capable AI planning systems that combine the flexibility of LLMs with the rigor of traditional control approaches.

\bibliographystyle{plainnat}
\bibliography{references}

\begin{thebibliography}{27}
\providecommand{\natexlab}[1]{#1}
\providecommand{\url}[1]{\texttt{#1}}
\expandafter\ifx\csname urlstyle\endcsname\relax
  \providecommand{\doi}[1]{doi: #1}\else
  \providecommand{\doi}{doi: \begingroup \urlstyle{rm}\Url}\fi

\bibitem[Ahn et~al.(2022)Ahn, Brohan, Brown, Chebotar, Cortes, David, Finn, Fu, Gopalakrishnan, Hausman, Herzog, Ho, Hsu, Ibarz, Ichter, Irpan, Jang, Ruano, Jeffrey, Jesmonth, Joshi, Julian, Kalashnikov, Kuang, Lee, Levine, Lu, Luu, Parada, Pastor, Quiambao, Rao, Rettinghouse, Reyes, Sermanet, Sievers, Tan, Toshev, Vanhoucke, Xia, Xiao, Xu, Xu, Yan, and Zeng]{ahn2022doasican}
Michael Ahn, Anthony Brohan, Noah Brown, Yevgen Chebotar, Omar Cortes, Byron David, Chelsea Finn, Chuyuan Fu, Keerthana Gopalakrishnan, Karol Hausman, Alex Herzog, Daniel Ho, Jasmine Hsu, Julian Ibarz, Brian Ichter, Alex Irpan, Eric Jang, Rosario~Jauregui Ruano, Kyle Jeffrey, Sally Jesmonth, Nikhil~J Joshi, Ryan Julian, Dmitry Kalashnikov, Yuheng Kuang, Kuang-Huei Lee, Sergey Levine, Yao Lu, Linda Luu, Carolina Parada, Peter Pastor, Jornell Quiambao, Kanishka Rao, Jarek Rettinghouse, Diego Reyes, Pierre Sermanet, Nicolas Sievers, Clayton Tan, Alexander Toshev, Vincent Vanhoucke, Fei Xia, Ted Xiao, Peng Xu, Sichun Xu, Mengyuan Yan, and Andy Zeng.
\newblock {Do As I Can, Not As I Say: Grounding Language in Robotic Affordances}.
\newblock \emph{arXiv preprint arXiv:2204.01691}, 2022.

\bibitem[Brown et~al.(2020)Brown, Mann, Ryder, Subbiah, Kaplan, Dhariwal, Neelakantan, Shyam, Sastry, Askell, Agarwal, Herbert-Voss, Krueger, Henighan, Child, Ramesh, Ziegler, Wu, Winter, Hesse, Chen, Sigler, Litwin, Gray, Chess, Clark, Berner, McCandlish, Radford, Sutskever, and Amodei]{NEURIPS2020_1457c0d6}
Tom Brown, Benjamin Mann, Nick Ryder, Melanie Subbiah, Jared~D Kaplan, Prafulla Dhariwal, Arvind Neelakantan, Pranav Shyam, Girish Sastry, Amanda Askell, Sandhini Agarwal, Ariel Herbert-Voss, Gretchen Krueger, Tom Henighan, Rewon Child, Aditya Ramesh, Daniel Ziegler, Jeffrey Wu, Clemens Winter, Chris Hesse, Mark Chen, Eric Sigler, Mateusz Litwin, Scott Gray, Benjamin Chess, Jack Clark, Christopher Berner, Sam McCandlish, Alec Radford, Ilya Sutskever, and Dario Amodei.
\newblock {Language Models are Few-Shot Learners}.
\newblock In H.~Larochelle, M.~Ranzato, R.~Hadsell, M.F. Balcan, and H.~Lin, editors, \emph{Advances in Neural Information Processing Systems}, volume~33, pages 1877--1901. Curran Associates, Inc., 2020.
\newblock URL \url{https://proceedings.neurips.cc/paper_files/paper/2020/file/1457c0d6bfcb4967418bfb8ac142f64a-Paper.pdf}.

\bibitem[Chae et~al.(2024)Chae, Kim, iunn Ong, Gwak, Song, Kim, Kim, Lee, and Yeo]{chae2024webagentsworldmodels}
Hyungjoo Chae, Namyoung Kim, Kai~Tzu iunn Ong, Minju Gwak, Gwanwoo Song, Jihoon Kim, Sunghwan Kim, Dongha Lee, and Jinyoung Yeo.
\newblock {Web Agents with World Models: Learning and Leveraging Environment Dynamics in Web Navigation}, 2024.
\newblock URL \url{https://arxiv.org/abs/2410.13232}.

\bibitem[Chen et~al.(2024)Chen, Pesaranghader, Sadhu, and Yi]{chen2024travelplanner}
Yanan Chen, Ali Pesaranghader, Tanmana Sadhu, and Dong~Hoon Yi.
\newblock Can we rely on llm agents to draft long-horizon plans? let's take travelplanner as an example.
\newblock \emph{arXiv preprint}, arXiv:2408.06318, 2024.
\newblock URL \url{https://doi.org/10.48550/arXiv.2408.06318}.

\bibitem[DeepSeek-AI et~al.(2025)DeepSeek-AI, Guo, Yang, Zhang, Song, Zhang, Xu, Zhu, Ma, Wang, Bi, Zhang, Yu, Wu, Wu, Gou, Shao, Li, Gao, Liu, Xue, Wang, Wu, Feng, Lu, Zhao, Deng, Zhang, Ruan, Dai, Chen, Ji, Li, Lin, Dai, Luo, Hao, Chen, Li, Zhang, Bao, Xu, Wang, Ding, Xin, Gao, Qu, Li, Guo, Li, Wang, Chen, Yuan, Qiu, Li, Cai, Ni, Liang, Chen, Dong, Hu, Gao, Guan, Huang, Yu, Wang, Zhang, Zhao, Wang, Zhang, Xu, Xia, Zhang, Zhang, Tang, Li, Wang, Li, Tian, Huang, Zhang, Wang, Chen, Du, Ge, Zhang, Pan, Wang, Chen, Jin, Chen, Lu, Zhou, Chen, Ye, Wang, Yu, Zhou, Pan, Li, et~al.]{DeepSeekR1_2025}
DeepSeek-AI, Daya Guo, Dejian Yang, Haowei Zhang, Junxiao Song, Ruoyu Zhang, Runxin Xu, Qihao Zhu, Shirong Ma, Peiyi Wang, Xiao Bi, Xiaokang Zhang, Xingkai Yu, Yu~Wu, Z.F. Wu, Zhibin Gou, Zhihong Shao, Zhuoshu Li, Ziyi Gao, Aixin Liu, Bing Xue, Bingxuan Wang, Bochao Wu, Bei Feng, Chengda Lu, Chenggang Zhao, Chengqi Deng, Chenyu Zhang, Chong Ruan, Damai Dai, Deli Chen, Dongjie Ji, Erhang Li, Fangyun Lin, Fucong Dai, Fuli Luo, Guangbo Hao, Guanting Chen, Guowei Li, H.~Zhang, Han Bao, Hanwei Xu, Haocheng Wang, Honghui Ding, Huajian Xin, Huazuo Gao, Hui Qu, Hui Li, Jianzhong Guo, Jiashi Li, Jiawei Wang, Jingchang Chen, Jingyang Yuan, Junjie Qiu, Junlong Li, J.L. Cai, Jiaqi Ni, Jian Liang, Jin Chen, Kai Dong, Kai Hu, Kaige Gao, Kang Guan, Kexin Huang, Kuai Yu, Lean Wang, Lecong Zhang, Liang Zhao, Litong Wang, Liyue Zhang, Lei Xu, Leyi Xia, Mingchuan Zhang, Minghua Zhang, Minghui Tang, Meng Li, Miaojun Wang, Mingming Li, Ning Tian, Panpan Huang, Peng Zhang, Qiancheng Wang, Qinyu Chen, Qiushi Du, Ruiqi Ge, Ruisong Zhang, Ruizhe Pan, Runji Wang, R.J. Chen, R.L. Jin, Ruyi Chen, Shanghao Lu, Shangyan Zhou, Shanhuang Chen, Shengfeng Ye, Shiyu Wang, Shuiping Yu, Shunfeng Zhou, Shuting Pan, S.S. Li, et~al.
\newblock {DeepSeek-R1: Incentivizing Reasoning Capability in LLMs via Reinforcement Learning}.
\newblock \url{https://doi.org/10.48550/arXiv.2501.12948}, Jan 2025.
\newblock arXiv:2501.12948 [cs.CL].

\bibitem[García et~al.(1989)García, Prett, and Morari]{GARCIA1989335}
Carlos~E. García, David~M. Prett, and Manfred Morari.
\newblock {Model predictive control: Theory and practice—A survey}.
\newblock \emph{Automatica}, 25\penalty0 (3):\penalty0 335--348, 1989.
\newblock ISSN 0005-1098.
\newblock \doi{https://doi.org/10.1016/0005-1098(89)90002-2}.
\newblock URL \url{https://www.sciencedirect.com/science/article/pii/0005109889900022}.

\bibitem[Gu et~al.(2024)Gu, Zheng, Gou, Zhang, Chang, Srivastava, Xie, Qi, Sun, and Su]{gu2024llmsecretlyworldmodel}
Yu~Gu, Boyuan Zheng, Boyu Gou, Kai Zhang, Cheng Chang, Sanjari Srivastava, Yanan Xie, Peng Qi, Huan Sun, and Yu~Su.
\newblock {Is Your LLM Secretly a World Model of the Internet? Model-Based Planning for Web Agents}, 2024.
\newblock URL \url{https://arxiv.org/abs/2411.06559}.

\bibitem[Hao et~al.(2023)Hao, Gu, Ma, Hong, Wang, Wang, and Hu]{hao2023reasoninglanguagemodelplanning}
Shibo Hao, Yi~Gu, Haodi Ma, Joshua~Jiahua Hong, Zhen Wang, Daisy~Zhe Wang, and Zhiting Hu.
\newblock Reasoning with language model is planning with world model, 2023.
\newblock URL \url{https://arxiv.org/abs/2305.14992}.

\bibitem[Holtzman et~al.(2020)Holtzman, Buys, Du, Forbes, and Choi]{holtzman2020curiouscaseneuraltext}
Ari Holtzman, Jan Buys, Li~Du, Maxwell Forbes, and Yejin Choi.
\newblock {The Curious Case of Neural Text Degeneration}, 2020.
\newblock URL \url{https://arxiv.org/abs/1904.09751}.

\bibitem[Jiang et~al.(2024)Jiang, Chen, Min, Chen, Cheng, Wang, Tang, Sun, Deng, Zhao, Liu, Yan, Xie, Wang, and Wen]{jiang2024treesearch}
Jinhao Jiang, Zhipeng Chen, Yingqian Min, Jie Chen, Xiaoxue Cheng, Jiapeng Wang, Yiru Tang, Haoxiang Sun, Jia Deng, Wayne~Xin Zhao, Zheng Liu, Dong Yan, Jian Xie, Zhongyuan Wang, and Ji-Rong Wen.
\newblock {Enhancing LLM Reasoning with Reward-guided Tree Search}.
\newblock \emph{arXiv preprint arXiv:2411.11694}, 2024.

\bibitem[Light et~al.(2024)Light, Wu, Sun, Yu, liu, Zhao, Hu, Chen, and Cheng]{light2024scatteredforestsearchsmarter}
Jonathan Light, Yue Wu, Yiyou Sun, Wenchao Yu, Yanchi liu, Xujiang Zhao, Ziniu Hu, Haifeng Chen, and Wei Cheng.
\newblock Scattered forest search: Smarter code space exploration with llms, 2024.
\newblock URL \url{https://arxiv.org/abs/2411.05010}.

\bibitem[Mayne et~al.(2000)Mayne, Rawlings, Rao, and Scokaert]{mayne2000mpc}
D.~Q. Mayne, J.~B. Rawlings, C.~V. Rao, and P.~O.~M. Scokaert.
\newblock {Constrained model predictive control: Stability and optimality}.
\newblock \emph{Automatica}, 36\penalty0 (6):\penalty0 789--814, 2000.

\bibitem[{Paa{\ss}, Gerhard and Giesselbach, Sven}(2023)]{Paaß2023}
{Paa{\ss}, Gerhard and Giesselbach, Sven}.
\newblock \emph{Pre-trained Language Models}, pages 19--78.
\newblock Springer International Publishing, Cham, 2023.
\newblock ISBN 978-3-031-23190-2.
\newblock \doi{10.1007/978-3-031-23190-2_2}.
\newblock URL \url{https://doi.org/10.1007/978-3-031-23190-2_2}.

\bibitem[Paul(2024)]{paul2024continualplanning}
Swarna~Kamal Paul.
\newblock {Continually Learning Planning Agent for Large Environments guided by LLMs}.
\newblock In \emph{2024 IEEE Conference on Artificial Intelligence (CAI)}, pages 377--382, 2024.
\newblock \doi{10.1109/CAI59869.2024.00076}.

\bibitem[Putta et~al.(2024)Putta, Mills, Garg, Motwani, Finn, Garg, and Rafailov]{putta2024agentq}
Pranav Putta, Edmund Mills, Naman Garg, Sumeet Motwani, Chelsea Finn, Divyansh Garg, and Rafael Rafailov.
\newblock {Agent Q: Advanced Reasoning and Learning for Autonomous AI Agents}.
\newblock \emph{arXiv preprint arXiv:2408.07199}, 2024.
\newblock \doi{10.48550/arXiv.2408.07199}.
\newblock Version 1, submitted on 13 Aug 2024.

\bibitem[Singh et~al.(2024)Singh, Traum, and Thomason]{singh2024twostepmultiagenttaskplanning}
Ishika Singh, David Traum, and Jesse Thomason.
\newblock {TwoStep: Multi-agent Task Planning using Classical Planners and Large Language Models}, 2024.
\newblock URL \url{https://arxiv.org/abs/2403.17246}.

\bibitem[Song et~al.(2023)Song, Wu, Washington, Sadler, Chao, and Su]{song2023llmplanner}
Chan~Hee Song, Jiaman Wu, Clayton Washington, Brian~M. Sadler, Wei-Lun Chao, and Yu~Su.
\newblock {LLM-Planner: Few-Shot Grounded Planning for Embodied Agents with Large Language Models}.
\newblock In \emph{Proceedings of the IEEE/CVF International Conference on Computer Vision (ICCV)}, pages 2998--3009, 2023.

\bibitem[Wang et~al.(2024{\natexlab{a}})Wang, Song, Tian, Peng, Yu, Mi, Su, and Yu]{wang2024litesearch}
Ante Wang, Linfeng Song, Ye~Tian, Baolin Peng, Dian Yu, Haitao Mi, Jingsong Su, and Dong Yu.
\newblock {LiteSearch: Efficacious Tree Search for LLM}.
\newblock \emph{arXiv preprint arXiv:2407.00320}, 2024{\natexlab{a}}.

\bibitem[Wang et~al.(2024{\natexlab{b}})Wang, Cassano, Wu, Bai, Song, Nath, Han, Hendryx, Yue, and Zhang]{wang2024planningnaturallanguageimproves}
Evan Wang, Federico Cassano, Catherine Wu, Yunfeng Bai, Will Song, Vaskar Nath, Ziwen Han, Sean Hendryx, Summer Yue, and Hugh Zhang.
\newblock Planning in natural language improves llm search for code generation, 2024{\natexlab{b}}.
\newblock URL \url{https://arxiv.org/abs/2409.03733}.

\bibitem[Wang et~al.(2023)Wang, Xie, Jiang, Mandlekar, Xiao, Zhu, Fan, and Anandkumar]{wang2023voyager}
Guanzhi Wang, Yuqi Xie, Yunfan Jiang, Ajay Mandlekar, Chaowei Xiao, Yuke Zhu, Linxi Fan, and Anima Anandkumar.
\newblock {Voyager: An Open-Ended Embodied Agent with Large Language Models}.
\newblock \emph{arXiv preprint arXiv:2305.16291}, 2023.

\bibitem[Wang et~al.(2022)Wang, Wei, Schuurmans, Le, Chi, Narang, Chowdhery, and Zhou]{wang2022selfconsistency}
Xuezhi Wang, Jason Wei, Dale Schuurmans, Quoc Le, Ed~Chi, Sharan Narang, Aakanksha Chowdhery, and Denny Zhou.
\newblock {Self-Consistency Improves Chain of Thought Reasoning in Language Models}.
\newblock \emph{arXiv preprint arXiv:2203.11171}, 2022.

\bibitem[Wei et~al.(2022)Wei, Wang, Schuurmans, Bosma, Ichter, Xia, Chi, Le, and Zhou]{wei2022chainofthought}
Jason Wei, Xuezhi Wang, Dale Schuurmans, Maarten Bosma, Brian Ichter, Fei Xia, Ed~Chi, Quoc Le, and Denny Zhou.
\newblock {Chain-of-Thought Prompting Elicits Reasoning in Large Language Models}.
\newblock \emph{arXiv preprint arXiv:2201.11903}, 2022.

\bibitem[Xiong et~al.(2024)Xiong, Payani, Yang, and Fekri]{xiong2024deliberatereasoningllmsstructureaware}
Siheng Xiong, Ali Payani, Yuan Yang, and Faramarz Fekri.
\newblock {Deliberate Reasoning for LLMs as Structure-aware Planning with Accurate World Model}, 2024.
\newblock URL \url{https://arxiv.org/abs/2410.03136}.

\bibitem[Zhang et~al.(2024)Zhang, Deik, Li, Zhang, and Liu]{zhang2024planningmulticonstraintscollaborativelanguage}
Cong Zhang, Derrick Goh~Xin Deik, Dexun Li, Hao Zhang, and Yong Liu.
\newblock {Planning with Multi-Constraints via Collaborative Language Agents}, 2024.
\newblock URL \url{https://arxiv.org/abs/2405.16510}.

\bibitem[Zheng et~al.(2025)Zheng, Mishra, Zhang, Chen, Chen, Nova, Hou, Cheng, Le, and Zhou]{zheng2024naturalplan}
Huaixiu~Steven Zheng, Swaroop Mishra, Hugh Zhang, Xinyu Chen, Minmin Chen, Azade Nova, Le~Hou, Heng-Tze Cheng, Quoc~V. Le, and Denny Zhou.
\newblock {Natural Plan: Benchmarking LLMs on Natural Language Planning}.
\newblock \emph{arXiv preprint arXiv:2406.04520}, 2025.

\bibitem[Zhou et~al.(2024)Zhou, Song, Yao, Shu, and Ma]{zhou2024isr}
Zhehua Zhou, Jiayang Song, Kunpeng Yao, Zhan Shu, and Lei Ma.
\newblock {ISR-LLM: Iterative Self-Refined Large Language Model for Long-Horizon Sequential Task Planning}.
\newblock In \emph{Proceedings of the 2024 IEEE International Conference on Robotics and Automation (ICRA)}, Yokohama, Japan, May 2024. IEEE.
\newblock \doi{10.1109/ICRA57147.2024.10610065}.

\bibitem[Zhu et~al.(2024)Zhu, Qiao, Ou, Deng, Zhang, Lyu, Shen, Liang, Gu, and Chen]{zhu2024knowagent}
Yuqi Zhu, Shuofei Qiao, Yixin Ou, Shumin Deng, Ningyu Zhang, Shiwei Lyu, Yue Shen, Lei Liang, Jinjie Gu, and Huajun Chen.
\newblock {KnowAgent: Knowledge-Augmented Planning for LLM-Based Agents}.
\newblock \emph{arXiv preprint arXiv:2403.03101}, 2024.
\newblock URL \url{https://doi.org/10.48550/arXiv.2403.03101}.

\end{thebibliography}

\appendix

\section{Mass-Spring System}
\begin{lstlisting}[caption={LLMPC prompt template for mass-spring problem}, label={lst:springprompt}]
    prompt = f"""
Given:
 - A mass-spring system with position x and velocity v.
 - Dynamics:
   x_(k+1) = x_k + dt * v_k
   v_(k+1) = v_k + (dt/m)*(u_k - k_spring*x_k)
 - Parameters: m={m}, k_spring={k_spring}, dt={dt}
 - Current state: x={x_init}, v={v_init}
 - Goal position: x_goal={x_goal}
 - Horizon: H={H}
 - The current spring force is {-k_spring*x_init}
 
 You control the force on the spring via the control sequence u = [u_0, u_1, ..., u_{H-1}].
 You must apply forces to get the spring to the goal position.
 Please propose {K} candidate control sequences, each being a list of length H.
 - Controls should be between 0 and 20
 - Return them as a Python dictionary with keys "sequence_1", "sequence_2", ..., "sequence_{K}", 
   where each value is a list of length H. Example:
   {{
     "sequence_1": [u_0, u_1, ..., u_{{H-1}}],
     "sequence_2": [u_0, u_1, ..., u_{{H-1}}],
     ...
   }}
Do not use ```python tags, no extra commentary, just return the dictionary.
"""
\end{lstlisting}

\section{Trip Planning}
\begin{lstlisting}[basicstyle=\tiny, caption={Example trip planning problem}, label={lst:trip_example}]
TASK: 
You plan to visit 10 European cities for 21 days in total. You only take direct flights to commute between cities. You plan to stay in Vienna for 3 days. You want to spend 5 days in Frankfurt. You want to spend 2 days in Oslo. You are going to attend a wedding in Oslo between day 20 and day 21. You want to spend 3 days in Prague. You would like to visit Valencia for 2 days. You want to meet a friend in Valencia between day 17 and day 18. You plan to stay in Dubrovnik for 2 days. You would like to visit Edinburgh for 5 days. From day 1 to day 5, there is a annual show you want to attend in Edinburgh. You plan to stay in London for 2 days. You plan to visit relatives in London between day 12 and day 13. You would like to visit Munich for 3 days. You would like to meet your friends at Munich between day 18 and day 20 to tour together. You would like to visit Budapest for 3 days.

Here are the cities that have direct flights:\nValencia and Munich, Vienna and Munich, Vienna and Valencia, London and Budapest, London and Oslo, Edinburgh and Budapest, Frankfurt and Budapest, Frankfurt and London, Prague and Oslo, Edinburgh and Oslo, Edinburgh and Munich, Prague and Munich, London and Prague, Edinburgh and London, Edinburgh and Frankfurt, Dubrovnik and Munich, Dubrovnik and Vienna, Munich and Oslo, Dubrovnik and Oslo, Budapest and Munich, Frankfurt and Prague, Vienna and London, Frankfurt and Vienna, Frankfurt and Oslo, Frankfurt and Munich, Vienna and Oslo, Vienna and Prague, Budapest and Oslo, Budapest and Prague, London and Valencia, London and Munich, Frankfurt and Dubrovnik, Prague and Valencia, Frankfurt and Valencia, Edinburgh and Prague, Vienna and Budapest.

Find a trip plan of visiting the cities for 21 days by taking direct flights to commute between them.\nSOLUTION: Here is the trip plan for visiting the 10 European cities for 21 days:

SOLUTION:
**Day 1-5:** Arriving in Edinburgh and visit Edinburgh for 5 days.
**Day 5:** Fly from Edinburgh to Frankfurt.
**Day 5-9:** Visit Frankfurt for 5 days.
**Day 9:** Fly from Frankfurt to Dubrovnik.
**Day 9-10:** Visit Dubrovnik for 2 days.
**Day 10:** Fly from Dubrovnik to Vienna.
**Day 10-12:** Visit Vienna for 3 days.
**Day 12:** Fly from Vienna to London.
**Day 12-13:** Visit London for 2 days.
**Day 13:** Fly from London to Budapest.
**Day 13-15:** Visit Budapest for 3 days.
**Day 15:** Fly from Budapest to Prague.
**Day 15-17:** Visit Prague for 3 days.
**Day 17:** Fly from Prague to Valencia.
**Day 17-18:** Visit Valencia for 2 days.
**Day 18:** Fly from Valencia to Munich.
**Day 18-20:** Visit Munich for 3 days.
**Day 20:** Fly from Munich to Oslo.
**Day 20-21:** Visit Oslo for 2 days.
\end{lstlisting}

\begin{lstlisting}[basicstyle=\tiny, caption={Trip Planning LLMPC System Prompt Template}, label={lst:trip_system}]
You are an expert travel planner assistant. Your goal is to create and refine travel plans that satisfy all given constraints.
Your only job is to focus on the constraints around cities to visit, number of days and ordering of the trip.
You do not need to investigate activities, accomodation etc, only focus on satisfying the stated trip constraints.
If you are revising an existing plan, don't forget to consider changing the starting city of the trip or reordering the cities visited.

Specifically you will be asked to propose a trip plan given constraints on the number of days, flights and order of locations to visit.
Note that when flying from one city to another it counts as a day spent in both cities and will count towards the number of days required to visit both of those cities. Take this into account when making your plan.
For example if we fly from city A to city B on day 7 the visit to city B will start on day 7 and the visit to city A will end on day 7.

Here are example Task descriptions and solutions:
TASK:
You plan to visit european cities for 10 days. You want to spend 5 days in Rome, 4 days in Amsterdam and 3 days in Paris.
You plan to meet a friend in Paris on the 9th day of the trip.
There are direct flights between Rome and Paris, Rome and Amsterdam.

Find a trip plan of visiting the cities for 10 days by taking direct flights to commute between them.

PLAN:
**Day 1-4:** Visit Amsterdam for 4 days.
**Day 4:** Fly from Rome to Amsterdam.
**Day 4-8:** Visit Rome for 5 days.
**Day 8:** Fly from Rome to Paris.
**Day 8-10:** Visit Paris for 3 days, spend the 9th day with your friend as planned.

TASK:
You have been asked to solve the following trip planning task:
You plan to visit 3 European cities for 15 days in total. You only take direct flights to commute between cities. You want to spend 6 days in Athens. You want to spend 4 days in London. You want to spend 7 days in Madrid.

Here are the cities that have direct flights:
Madrid and London, from London to Athens.

Find a trip plan of visiting the cities for 15 days by taking direct flights to commute between them.

Output:
**Day 1-7:** Visit Madrid for 7 days.
**Day 7:** Fly from Madrid to London.
**Day 7-10:** Visit London for 4 days.
**Day 10:** Fly from London to Athens.
**Day 10-15:** Visit Athens for 6 days.

TASK:
You plan to visit 8 European cities for 22 days in total. You only take direct flights to commute between cities. You would like to visit Vilnius for 4 days. You would like to visit Venice for 5 days. You plan to stay in Warsaw for 4 days. You want to meet a friend in Warsaw between day 14 and day 17. You want to spend 5 days in Mykonos. You plan to stay in Salzburg for 5 days. You plan to stay in Amsterdam for 2 days. You would like to meet your friends at Amsterdam between day 17 and day 18 to tour together. You plan to stay in Hamburg for 2 days. You would like to visit Copenhagen for 2 days.

Here are the cities that have direct flights:
Warsaw and Amsterdam, Hamburg and Venice, Hamburg and Warsaw, Venice and Warsaw, Hamburg and Amsterdam, Venice and Copenhagen, Vilnius and Amsterdam, Vilnius and Warsaw, Hamburg and Copenhagen, Salzburg and Hamburg, Copenhagen and Amsterdam, Copenhagen and Vilnius, Copenhagen and Warsaw, Venice and Amsterdam, Amsterdam and Mykonos.

Find a trip plan of visiting the cities for 22 days by taking direct flights to commute between them.

Output:
**Day 1-5:** Visit Salzburg for 5 days.
**Day 5:** Fly from Salzburg to Hamburg.
**Day 5-6:** Visit Hamburg for 2 days.
**Day 6:** Fly from Hamburg to Venice.
**Day 6-10:** Visit Venice for 5 days.
**Day 10:** Fly from Venice to Copenhagen.
**Day 10-11:** Visit Copenhagen for 2 days.
**Day 11:** Fly from Copenhagen to Vilnius.
**Day 11-14:** Visit Vilnius for 4 days.
**Day 14:** Fly from Vilnius to Warsaw.
**Day 14-17:** Visit Warsaw for 4 days.
**Day 17:** Fly from Warsaw to Amsterdam.
**Day 17-18:** Visit Amsterdam for 2 days.
**Day 18:** Fly from Amsterdam to Mykonos.
**Day 18-22**: Visit Mykonos for 5 days.

Always format your response using the following template:

PLAN:
<your complete trip plan here>
\end{lstlisting}

\begin{lstlisting}[basicstyle=\tiny, caption={Trip Planning LLMPC Instruction Prompt Template}, label={lst:trip_instruction}]
You have been asked to solve the following trip planning task:
TASK:
{task}

Your current trip plan is:
{current_plan}

{feedback_string}
\end{lstlisting}

\section{Meeting Planning}

\begin{lstlisting}[basicstyle=\tiny, caption={Example meeting planning problem}, label={lst:meeting_example}]
You are visiting San Francisco for the day and want to meet as many friends as possible. Solve the problem by considering various different schedules and picking the best one to optimize your goals.

Travel distances (in minutes):
Marina District to Alamo Square: 15.
Marina District to Fisherman's Wharf: 10.
Marina District to Union Square: 16.
Marina District to Embarcadero: 14.
Marina District to Financial District: 17.
Alamo Square to Marina District: 15.
Alamo Square to Fisherman's Wharf: 19.
Alamo Square to Union Square: 14.
Alamo Square to Embarcadero: 17.
Alamo Square to Financial District: 17.
Fisherman's Wharf to Marina District: 9.
Fisherman's Wharf to Alamo Square: 20.
Fisherman's Wharf to Union Square: 13.
Fisherman's Wharf to Embarcadero: 8.
Fisherman's Wharf to Financial District: 11.
Union Square to Marina District: 18.
Union Square to Alamo Square: 15.
Union Square to Fisherman's Wharf: 15.
Union Square to Embarcadero: 11.
Union Square to Financial District: 9.
Embarcadero to Marina District: 12.
Embarcadero to Alamo Square: 19.
Embarcadero to Fisherman's Wharf: 6.
Embarcadero to Union Square: 10.
Embarcadero to Financial District: 5.
Financial District to Marina District: 15.
Financial District to Alamo Square: 17.
Financial District to Fisherman's Wharf: 10.
Financial District to Union Square: 9.
Financial District to Embarcadero: 4.

CONSTRAINTS: You arrive at Marina District at 9:00AM. Mary will be at Alamo Square from 3:15PM to 6:30PM. You'd like to meet Mary for a minimum of 60 minutes. Deborah will be at Fisherman's Wharf from 7:00PM to 10:00PM. You'd like to meet Deborah for a minimum of 45 minutes. Jason will be at Union Square from 11:00AM to 1:15PM. You'd like to meet Jason for a minimum of 75 minutes. Betty will be at Embarcadero from 2:00PM to 6:15PM. You'd like to meet Betty for a minimum of 90 minutes. Anthony will be at Financial District from 12:15PM to 9:30PM. You'd like to meet Anthony for a minimum of 105 minutes.

SOLUTION:You start at Marina District at 9:00AM. You travel to Union Square in 16 minutes and arrive at 9:16AM. You wait until 11:00AM. You meet Jason for 75 minutes from 11:00AM to 12:15PM. You travel to Alamo Square in 15 minutes and arrive at 12:30PM. You wait until 3:15PM. You meet Mary for 60 minutes from 3:15PM to 4:15PM. You travel to Embarcadero in 17 minutes and arrive at 4:32PM. You meet Betty for 90 minutes from 4:32PM to 6:02PM. You travel to Financial District in 5 minutes and arrive at 6:07PM. You meet Anthony for 105 minutes from 6:07PM to 7:52PM. You travel to Fisherman's Wharf in 10 minutes and arrive at 8:02PM. You meet Deborah for 45 minutes from 8:02PM to 8:47PM.
\end{lstlisting}

\begin{lstlisting}[basicstyle=\tiny, caption={Meeting Planning LLMPC System Prompt Template}, label={lst:meeting_system}]
You are an expert meeting planner assistant. Your goal is to create and refine plans to meet friends at different places in the city, taking into account travel times and meeting time constraints.
Your job is to iteratively create and modify a plan that meets with all the listed friends for the duration and location specified in the constraints. 

You will be given a limited number of iteration to make a full plan, these will be indicated by STEP X/TOTAL_STEPS.
For each step, propose {PLANS_PER_ITERATION} different possible plans. Make the plans meaningfully different from each other (e.g. meeting friends in different orders, starting with different meetings).

When no plan is given you should start by planning meetings with the first few required people.
When an existing plan is given you should consider:
1. Adding more meetings
2. Modifying meeting orders
3. Adjusting meeting durations
4. Trying completely different meeting sequences

If you believe you have the best possible plan make no modifications and output the existing plan {PLANS_PER_ITERATION} times.
Do not propose meeting with imaginary friends. 
Only propose meetings with friends mentioned in the task description.
Do not propose multiple meetings with the same friend.
If the plan already includes all mentioned friends, do not make any modifications or propose additional meetings.

PLAN FORMAT:
A plan includes a start, meeting, traveling and waiting steps.
* The start of a plan should be phrased as 'You start at <location> at <start_time>'
* A meeting step should be phrased as 'You meet <friend_name> for <time_spent> minutes from <start_time> to <end_time>.'
* A travel step should be phrased as 'You travel to <location> in <travel_time> minutes and arrive at <arrival_time>.'
* A waiting step should be phrased as 'You wait until <end_time>.'

Only use the above phrasing, e.g. do not mention 'You travel back to...' only mention 'You travel to...'

EXAMPLES:
Here are example input task descriptions and output plans:

You are visiting San Francisco for the day and want to meet as many friends as possible. Solve the problem by considering various different schedules and picking the best one to optimize your goals.

Travel distances (in minutes):
Marina District to Alamo Square: 15.
Marina District to Fisherman's Wharf: 10.
Marina District to Union Square: 16.
Marina District to Embarcadero: 14.
Marina District to Financial District: 17.
Marina District to Nob Hill: 12.
Alamo Square to Marina District: 15.
Alamo Square to Fisherman's Wharf: 19.
Alamo Square to Union Square: 14.
Alamo Square to Embarcadero: 17.
Alamo Square to Financial District: 17.
Alamo Square to Nob Hill: 11.
Fisherman's Wharf to Marina District: 9.
Fisherman's Wharf to Alamo Square: 20.
Fisherman's Wharf to Union Square: 13.
Fisherman's Wharf to Embarcadero: 8.
Fisherman's Wharf to Financial District: 11.
Fisherman's Wharf to Nob Hill: 11.
Union Square to Marina District: 18.
Union Square to Alamo Square: 15.
Union Square to Fisherman's Wharf: 15.
Union Square to Embarcadero: 11.
Union Square to Financial District: 9.
Union Square to Nob Hill: 9.
Embarcadero to Marina District: 12.
Embarcadero to Alamo Square: 19.
Embarcadero to Fisherman's Wharf: 6.
Embarcadero to Union Square: 10.
Embarcadero to Financial District: 5.
Embarcadero to Nob Hill: 10.
Financial District to Marina District: 15.
Financial District to Alamo Square: 17.
Financial District to Fisherman's Wharf: 10.
Financial District to Union Square: 9.
Financial District to Embarcadero: 4.
Financial District to Nob Hill: 8.
Nob Hill to Marina District: 11.
Nob Hill to Alamo Square: 11.
Nob Hill to Fisherman's Wharf: 11.
Nob Hill to Union Square: 7.
Nob Hill to Embarcadero: 9.
Nob Hill to Financial District: 9.

CONSTRAINTS: You arrive at Marina District at 9:00AM. Deborah will be at Alamo Square from 11:15AM to 1:30PM. You'd like to meet Deborah for a minimum of 45 minutes. Jason will be at Fisherman's Wharf from 11:00AM to 1:15PM. You'd like to meet Jason for a minimum of 75 minutes. Betty will be at Union Square from 2:00PM to 6:15PM. You'd like to meet Betty for a minimum of 90 minutes. Anthony will be at Embarcadero from 12:15PM to 9:30PM. You'd like to meet Anthony for a minimum of 105 minutes. Daniel will be at Financial District from 7:00AM to 10:15AM. You'd like to meet Daniel for a minimum of 120 minutes. Jessica will be at Nob Hill from 6:00PM to 10:00PM. You'd like to meet Jessica for a minimum of 105 minutes.

SOLUTION:You start at Marina District at 9:00AM. You travel to Fisherman's Wharf in 10 minutes and arrive at 9:10AM. You wait until 11:00AM. You meet Jason for 75 minutes from 11:00AM to 12:15PM. You travel to Alamo Square in 20 minutes and arrive at 12:35PM. You meet Deborah for 45 minutes from 12:35PM to 1:20PM. You travel to Union Square in 14 minutes and arrive at 1:34PM. You wait until 2:00PM. You meet Betty for 90 minutes from 2:00PM to 3:30PM. You travel to Embarcadero in 11 minutes and arrive at 3:41PM. You meet Anthony for 105 minutes from 3:41PM to 5:26PM. You travel to Nob Hill in 10 minutes and arrive at 5:36PM. You wait until 6:00PM. You meet Jessica for 105 minutes from 6:00PM to 7:45PM.

OUTPUT FORMAT:

For your response only output the keyword SOLUTION followed by {PLANS_PER_ITERATION} plans separated by '---', do not number the plans or add headers, only output the text of the {PLANS_PER_ITERATION} plans, make sure the plans are different from each other, do not output anything other than this format:

SOLUTION:
<insert first plan>
---
<insert second plan>
---
...
\end{lstlisting}

\begin{lstlisting}[basicstyle=\tiny, caption={Meeting Planning LLMPC Instruction Prompt Template}, label={lst:meeting_instruction}]
STEP {step}/{total_steps}

You have been asked to solve the following meeting planning task, pay particular attention the constraints and propose 3 different possible plans according to the format described in the system prompt:
TASK:
{task}

Your current best meeting plan is:
{current_plan}

{feedback_string}

OUTPUT FORMAT:

For your response only output the keyword SOLUTION followed by {num_plans} plans separated by '---', do not number the plans or add headers, only output the text of the {num_plans} plans, make sure the plans are different from each other, do not output anything other than this format:

SOLUTION:
<insert first plan>
---
<insert second plan>
---
...
\end{lstlisting}

\end{document}